\documentclass[preprint,review,12pt]{elsarticle}

\usepackage{amssymb}

\usepackage{lineno}

\usepackage[hidelinks]{hyperref} 
\usepackage{booktabs} 
\usepackage{multirow} 
\usepackage{graphicx} 
\usepackage{amsmath} 
\usepackage{upgreek} 
\usepackage{xcolor}

\begin{document}

\begin{frontmatter}



\title{Explainable AI for Correct Root Cause Analysis of Product Quality in Injection Moulding}


\author[affil-flair]{Muhammad Muaz}
\author[affil-flair]{Sameed Sajid}
\author[affil-wzl]{Tobias Schulze}
\author[affil-flair]{Chang Liu}
\author[affil-wzl]{Nils Klasen}
\author[affil-flair]{Benny Drescher \corref{c1}}
\cortext[c1]{Email: \href{mailto:benny.drescher@hkflair.org}{benny.drescher@hkflair.org} (corresponding author) }
\affiliation[affil-flair]{organization={Hong Kong Industrial Artificial Intelligence and Robotics Centre (FLAIR)},
            addressline={Hong Kong Science Park, Shatin, NT}, 
            city={Hong Kong S.A.R.},
        	country={China}}
\affiliation[affil-wzl]{organization={Laboratory for Machine Tools and Production Engineering, WZL of RWTH Aachen University},
	addressline={Campus-Boulevard 30, 52074}, 
	city={Aachen},
	country={Germany}}

\begin{abstract}
If a product deviates from its desired properties in the injection moulding process, its root cause analysis can be aided by models that relate the input machine settings with the output quality characteristics. The machine learning models tested in the quality prediction are mostly black boxes; therefore, no direct explanation of their prognosis is given, which restricts their applicability in the quality control. The previously attempted explainability methods are either restricted to tree-based algorithms only or do not emphasize on the fact that some explainability methods can lead to wrong root cause identification of a product's deviation from its desired properties. This study first shows that the interactions among the multiple input machine settings do exist in real experimental data collected as per a central composite design. Then, the model-agnostic explainable AI methods are compared for the first time to show that different explainability methods indeed lead to different feature impact analysis in injection moulding. Moreover, it is shown that the better feature attribution translates to the correct cause identification and actionable insights for the injection moulding process. Being model agnostic, explanations on both random forest and multilayer perceptron are performed for the cause analysis, as both models have the mean absolute percentage error of less than 0.05\% on the experimental dataset.

\end{abstract}


%
%

\begin{keyword}
injection moulding \sep explainable artificial intelligence \sep quality control \sep predictive maintenance \sep deep learning 

\PACS 89.20.Bb \sep 07.05.Mh \sep 02.50.Le

\MSC[2010] 62P30 \sep 62M45 \sep 91A80

\end{keyword}

\end{frontmatter}


\section{Introduction}
\label{sec_intro}
Predicting a product's quality along with automating the root cause analysis in injection moulding can enhance the consistency and reliability of produced products, enable operations in real-time with traceability, and mitigate the plastic scraps.

In the last two decades, many machine learning pipelines have been shown to model the input process parameters and/or machine settings with the quality characteristics of the output product e.g., support vector machine \cite{ribeiro_support_2005, song_optimization_2020}; fully connected neural networks \cite{lockner_induced_2021, lockner_transfer_2022, chen_artificial_2020, ke_quality_2020}; tree-based algorithms \cite{jung_application_2021, finkeldey_learning_2020}; autoencoders \cite{jung_application_2021}; $ k $-nearest neighbour and naïve Bayes \cite{parizs_machine_2022}. The establishment of IoT alongside the Manufacturing Executive System (MES) for controlling machine peripherals, such as hot runner controllers, hopper dryers, coolant temperature, flow rate, and robotic de-moulding, has allowed for precise control over part quality by uploading data to the MES. As the data is increasingly made available, the injection moulding process in real-time can be captured and modelled more comprehensively (e.g., \cite{muaz_multitask_2023}). Most of the complex machine learning models are black boxes, which makes it difficult to understand the cause of a certain quality outcome. Finding such a cause is important to aid in the root cause analysis of undesired products. Letting the model speak about which parameter is the dominant cause and how to tweak it can help to automate not just the quality control process but also the fine tuning of machine settings to a new machine/mould/product.

For a human to be able to attribute the cause of a certain machine learning model's output among the inputs is referred to as explainability.  Some simple machine learning models are intrinsically interpretable/explainable such as linear regression, decision tree, and generalized additive models but these also get harder to interpret for humans when the model has many correlated input parameters. Nonetheless, more sophisticated models have been suggested for industrial productions with tabular datasets such as ensemble learning \cite{quatrini_machine_2020} and fully connected neural networks \cite{lockner_transfer_2022} due to their ability to capture the complex relationships and feature interactions. These sophisticated models, however, comes with the drawback of no inherent interpretability/explainability. To explain these models, there exists explainability methods, where only the input data and the prediction of the model are needed to provide the explanation \cite{vilone_classification_2021, dosilovic_explainable_2018, adadi_peeking_2018}. 

\begin{figure}[h]
	\centering
	\includegraphics[scale=0.5]{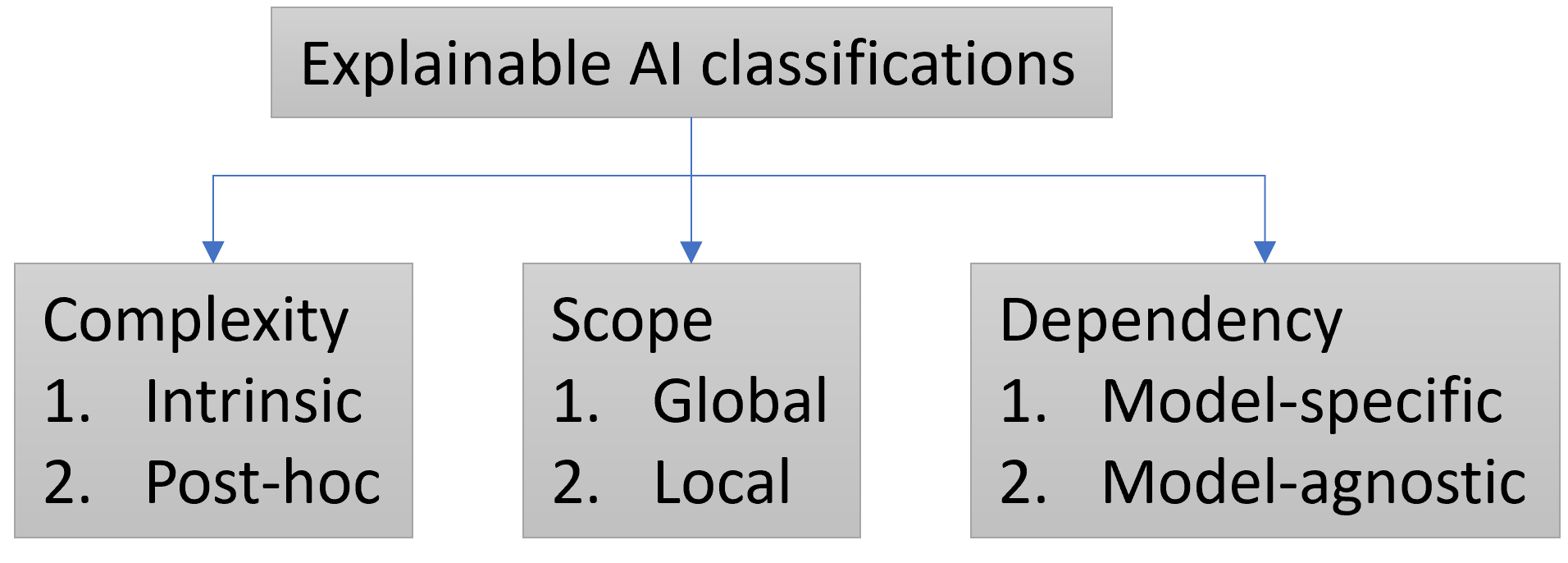}
	\caption{Explainable AI broad categories used in manufacturing \cite{dam_explainable_2018}.} 
	\label{fig_xai_classes}
\end{figure}

Explainable AI methods have been widely adopted in manufacturing and are categorized into different types as shown in Figure \ref{fig_xai_classes} \cite{soldatos_review_2021}. 
Global methods infer the structure of the ML model altogether, while local methods provide transparency for individual decisions \cite{dam_explainable_2018}. A special kind are model-agnostic methods as they can be applied regardless of the underlying ML model.

A global, post-hoc, and model-agnostic explainable AI method adapted in manufacturing is partial dependence plot (PDP) \cite{obregon_rule-based_2021}. However, for explaining the root cause of individual product qualities in the experiments of this study, global methods are infeasible because they can only explain the overall average behaviour of a model. In explaining an individual sample outcome from a single production cycle, the local interpretability methods are suitable. The local post-hoc model-agnostic methods adopted in injection moulding are individual conditional expectation (ICE) plots \cite{soler_advancing_2023}, rule-based explanations, that are specific to tree-based models \cite{obregon_rule-based_2021}; and SHapley Additive exPlanations (SHAP) \cite{gim_interpretation_2023}.

Previous attempts have been made to interpret prediction models in cyclic manufacturing processes. For instance, \cite{kozjek_interpretative_2017, kozjek_data_2019} suggested the use of intrinsically interpretable decision tree based algorithms. \cite{obregon_rule-based_2021} used a combination of interpretation methods called rule-based explanations that involved feature importance ranking, rule simplifications, and finally, visualization of results with PDP and ICE. However, the rule-based explanation method is specific to tree-based algorithms. Additionally, it is not possible to look at interactions of more than two features simultaneously by utilizing the ICE plots. 

\cite{gim_interpretation_2023} used SHAP to explain the influence of section-wise features from transient process data in injection moulding. 
\cite{gim_-mold_2024} employed SHAP to explain the effects of in-mould conditions on part quality and optimizes them to address the heavy reliance on domain expertise. 
\cite{soler_advancing_2023} utilized Tree SHAP on XGBoost as the global method by using its summary plot and ICE as the local method.
However, SHAP and ICE may end up suggesting different root causes for an outcome and which of these will lead to the correct cause is not studied and unclear.

In this study, interactions among the operator-controlled machine settings in injection moulding are first shown to exist with the help of Friedman's H-statistic \cite{friedmanPredictiveLearningRule2008}. Then, a permutation-based SHAP is compared with ICE to show for the first time that the interactions among the machine settings make the root cause analysis different in the quality control of injection moulding. With the consideration of feature interactions in calculating the relationship scores, the misleading root cause of a certain product quality in injection moulding can be avoided. The models implemented on the dataset are random forest and multilayer perceptron, which are both intrinsically not explainable and are suitable for quality prediction in injection moulding \cite{lockner_transfer_2022, obregon_rule-based_2021, gim_-mold_2024}. One can opt for more state-of-the-art models that are also tested and reported in Section \ref{sec_results}, however, random forest and multilayer perceptron are later shown in the paper (Table \ref{tab_mae}) to have adequate fitting to the complexity of data in hand. Two accurate models are chosen to ensure the explanations are model agnostic. If one of the models is not accurate, the explanation emerging from that model will be different from the other one, invalidating the choice of models.

The rest of the paper is organized as follows: Section \ref{sec_setup} describes the manufacturing setup where the dataset is acquired and deployed. The chosen black-box models, the machine setting interactions, the theory of interpretation methods, and the comparison method of cause analysis are discussed in Section \ref{sec_methodology}. The results of the machine setting interactions and discussion on cause analysis by the model's interpretation are discussed in Section \ref{sec_results}. In the last Section \ref{sec_concl}, the paper is concluded.

\section{Production setup}
\label{sec_setup}

The Dataset is collected on Sumitomo (SHI) Demag Plastics Machinery IntElect 100-250. The utilized material is Sabic PP 579S. After a moulded part is ejected, the product is transferred to a digital scale Sartorius AG Signum 3 for weight measurement. Table \ref{tab_setup} reports further details about the specifications of the machine during the data collection i.e., clamping force, screw diameter, etc. 

The experiment is designed with Central Composite Design (CCD) having 6 machine settings (the model input parameters/features). 
There are $ 2^6 = 64 $ factorial points (cube points), $ 2 \times 6 = 12 $ axial points (star points), and 1 centre point. It is a face-centered CCD with $ \alpha $ equals to 1.
The 6 machine settings are the cooling time, injection speed, melt temperature, mould temperature, packing pressure, and packing time, which are usually considered important in impacting the product's quality characteristics \cite{lockner_transfer_2022, finkeldey_learning_2020, tsaiInverseModelInjection2017}. For the quality characteristics (the model output parameters), the product weight is considered. Several other process parameters as well as time-series pressure and temperature sensor's data from the machine are also acquired. However, they are not included in the modelling for simplicity as this study focuses on analyzing factors controllable by a machine operator, ensuring that when an interpretation method identifies the main cause of a product's deviation, it can be achieved practically.

\begin{table}[h]
	\caption{Data collection setup.}
	\label{tab_setup}
	\resizebox{\columnwidth}{!}{%
		\begin{tabular}{@{}lp{8cm}p{8cm}@{}}
			\toprule
			Hardware/Equipment Name & Model & Further Characteristics \\
			\midrule
			Injection moulding machine & Sumitomo (SHI) Demag Plastics Machinery IntElect 100-250 & Clamping force: 100 t; screw diameter: 30 mm; max metered volume: 113 cm$^3$; max mould height: 430 mm; min mould height: 200 nm; max volume flow: 212 cm$^3$; max injection pressure: 2800 bar \\
			Temperature control unit & HB-Therm AG, St. Gallen, Switzerland Series4 HB-140U1 & Temperature control medium: water; max temperature control medium temperature: 140$^{\circ}$C; volume flow: 1.5 l/min; pressure: 16 bar \\
			Coordinate measuring machine & Carl Zeiss QEC O-Inspect 442 &  \\
			Weight scale & Sartorius AG Signum 3 &  \\
			\bottomrule
		\end{tabular}
	}
\end{table}

\begin{table}[h]
	\caption{Range {[}min, max{]} of 6 input machine settings.}
	\label{tab_inputs}
	\resizebox{\textwidth}{!}{%
		\begin{tabular}{llllll}
			\toprule
			\begin{tabular}[c]{@{}l@{}}Injection speed \\ ($ {\rm cm^3/s} $)\end{tabular} & \begin{tabular}[c]{@{}l@{}}Cooling time \\ (s)\end{tabular} & \begin{tabular}[c]{@{}l@{}}Packing pressure \\ (bar)\end{tabular} & \begin{tabular}[c]{@{}l@{}}Packing time \\ (sec)\end{tabular} & \begin{tabular}[c]{@{}l@{}}Mould temperature \\ ($ {\rm ^{\circ} C} $)\end{tabular} & \begin{tabular}[c]{@{}l@{}}Melt temperature \\ ($ {\rm ^{\circ} C} $)\end{tabular} \\ \hline
			{[}11.71, 23.21{]} & {[}3.785, 7.785{]} & {[}250, 550{]} & {[}1.025, 3.025{]} & {[}30, 60{]} & {[}220, 260{]} \\ \bottomrule
		\end{tabular}
	}
\end{table} 

In total, 77 different combinations of the machine settings are run, each repeated 20 times resulting in 1540 product cycles. The sequence of processing with the defined machine setting combinations in the CCD is adapted to save time e.g., by clustering same temperatures for the mould and melt. The machine data is extracted via the machine control system as comma-separated values. The weight measurements are noted manually in a spreadsheet file. The machine and weight data are then combined via the process cycle numbers marked on the products.

This data is split using stratified random sampling into training (60\%), validation (10\%), and testing (30\%) sets. The model interpretation methods are applied on the testing sets.

\begin{figure}[h] 
	\centering
	\includegraphics[scale=0.78]{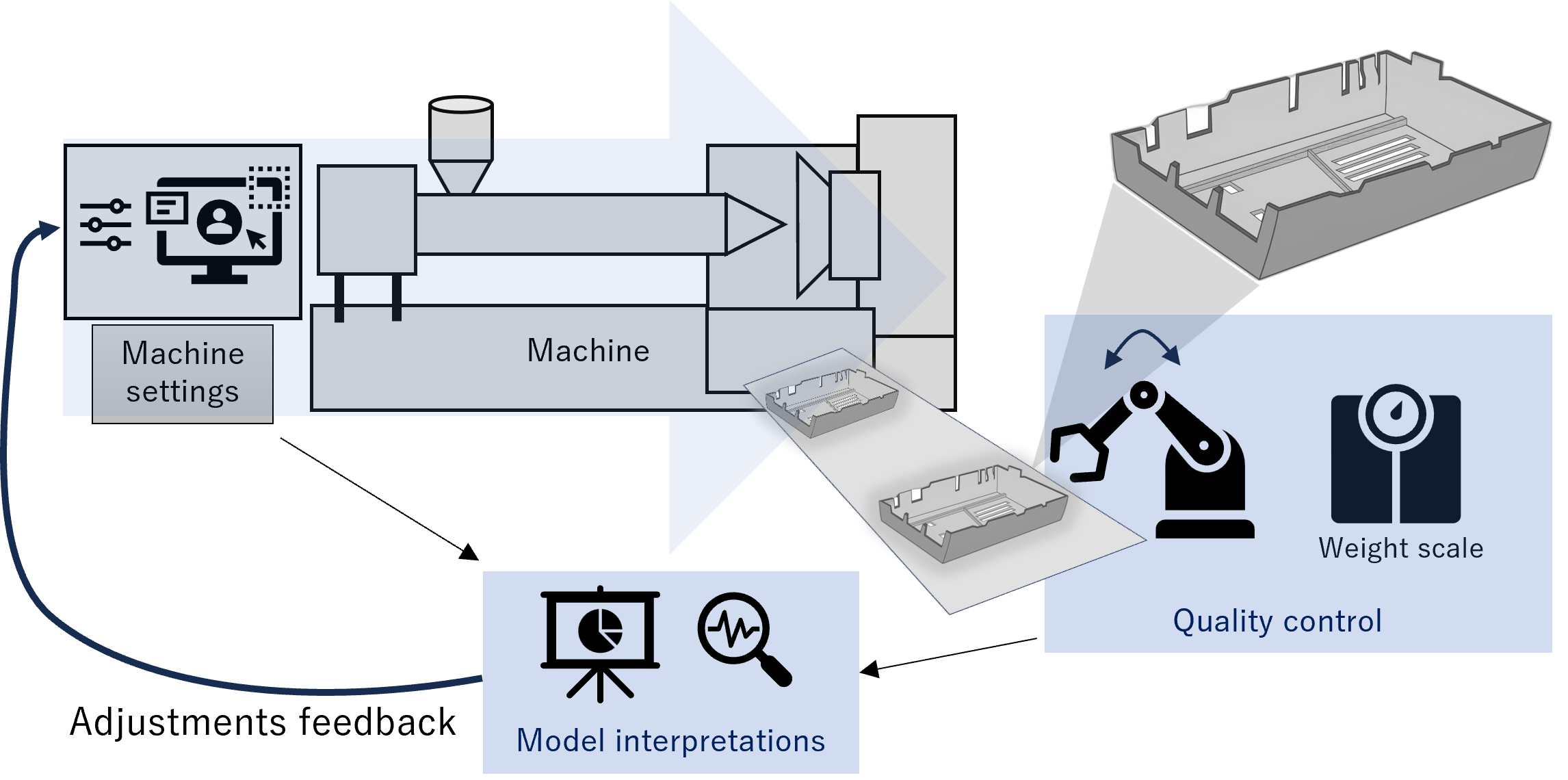}
	\caption{Schematic of production setup for training and deployment.}
	\label{fig_production_setup}
\end{figure}

After obtaining a process window for machine settings in injection moulding (e.g., through software like Autodesk Moldflow \cite{noauthor_moldflow_nodate}), a design of experiment provides with the plan to produce parts and assess the product quality in each cycle. Once the data is collected, it can be used to train a model that can capture the feature interactions comprehensively. This model is then explained and used for cause analysis to optimize the machine settings for obtaining the desirable quality characteristics. A schematic of the production environment visualizing the workflow is shown in Figure \ref{fig_production_setup}. 
After running on an initial combination of machine settings, the product is placed on the weight scale through a robotic arm. The machine settings along with the measured weight are provided to the explainer object, which then learns the impact of machine settings on the product weight, making it able to conduct the cause analysis during deployment. Based on the cause analysis, the new machine settings are recommended. A schematic of the actual product used in the experiments can also be seen in Figure \ref{fig_production_setup}.

\section{Methodology}
\label{sec_methodology}

\subsection{Chosen Models}
\label{subsec_meth_models}
Two popular black-box regression models are fitted and explained later by the interpretation methods: random forest (RF) and multilayer perceptron (MLP). RF is an ensemble of decision trees and uses bootstrap aggregation (sampling with replacement). A group of training sets $ D_m={(\mathbf{X}_m,\ \mathbf{y}_m\ )\ \ |\ \mathbf{X}_m,\mathbf{y}_m\in\mathfrak{R},\ m=1,\ 2,\ \ldots,M} $ is generated from the available data, and $ M $ decision trees are trained on them. Then, predictions from the individual decision trees applied on a test sample $ \mathbf{x}^\prime $ are combined as
\begin{equation} 
	\hat{y}=\frac{1}{M}\sum_{m=1}^{M}{y_m\left(\mathbf{x}^\prime\right)},
	\label{eq_rf}
\end{equation}
where $ \hat{y} $ denotes the predicted output. 
MLP is the feedforward neural network composed of fully connected layers and trains with backpropagation. MLP has strong capability to identify complex, inherently non-linear relationships. 

The hyperparameter selection process follows a structured, multi-stage approach to balance model performance and computational efficiency. A grid search is conducted over network architectures, selecting a shallow (8,4) configuration based on its stability and capacity for injection molding datasets. The learning rate ($ \eta ~=~ $ 0.01) is determined empirically to minimize loss oscillations, and regularization strength is tuned to control overfitting without degrading the predictive performance.
A more complex architecture can lead to optimization problems when applied to injection moulding, and a simple one worked well across a majority of injection moulding cases \cite{leeStudyArchitectureArtificial2023}. Table \ref{tab_mlp_hyperparameters} shows the hyperparameter values used when training the neural network. For the solver, the limited-memory BFGS (L-BFGS) gave lower mean absolute error as compared to Adam. 
For MLP, the data is subjected to standard scaling, to neutralize the difference in magnitudes among the different machine settings, ensuring the influence of the scale is consistent.

\begin{table}[h]
	\centering
	\caption{Hyperparameters of the MLP model.}
	\label{tab_mlp_hyperparameters}
	\begin{tabular}{@{}ll@{}}
		\toprule
		Hyperparameter & Value \\ \hline
		Number of hidden layers & 2 \\
		Number of neurons per layer & 8, 4 \\
		Batch size & 8 \\
		Maximum iterations & 200 \\
		Activation function & tanh \\
		Solver & L-BFGS \\
		Random state & 42 \\
		$ \text{L}_2 $ regularization & 0.0001 \\ \bottomrule
	\end{tabular}
\end{table}

Both RF and MLP have been used in the literature (e.g., see \cite{lockner_transfer_2022, finkeldey_learning_2020}) for quality prediction in injection moulding, however, both models are complex in the sense that the contribution of the input features in obtaining a particular quality of the product are not explainable by human.

The models' error performance and the way the data is split are important because only those trained models that are the correct representation of the data can be used to verify the explainability methods in this study. If the error performance with respect to the scale of the data is not satisfactory, the explanations obtained on top of such models should not be trusted and may lead to misleading insights. Both the mean absolute error (MAE) and mean absolute percentage error (MAPE) are reported for the error performance of the models. The collected data is precise and very little changes in the product weight (of the order of 0.1 grams) are significant with respect to the data's dynamic range. Therefore, MAE with respect to the data's dynamic range (named as range-scaled MAE (RSMAE)) is also reported (Equation $\left(\ref{eq_rsmae}\right)$).

\begin{equation} 
	\mathrm{RSMAE}~ := ~ \frac{\mathrm{MAE}}{\max{\left(y\right)}-\min{\left(y\right)}}\times100\ \% .
	\label{eq_rsmae}
\end{equation}

\subsection{Capturing Machine Setting Interactions}
\label{subsec_meth_interactions}
Interaction effects occur when the influence of one variable on the outcome depends on the values of other features. If features interact with one another, the prediction cannot be expressed as the sum of the feature effects \cite{molnar_interpretable_2022}. H-statistic \cite{friedmanPredictiveLearningRule2008, molnar_interpretable_2022} is used to check the existence of interaction among the machine settings in the experiments of this study.

By means of partial dependence, H-statistic quantifies the extent to which two features, $ x_j $ and $ x_k $  are interacting with each other, and it can also be used to measure the interaction of a feature with all other features in a model \cite{friedmanPredictiveLearningRule2008}. It is calculated by comparing the observed partial dependence function to another baseline version that assumes no interactions. The ratio of the variance explained by the interaction term to the total variance of the partial dependence function yields the H-statistic, which ranges from 0 (no interaction) to 1 or higher (strong interaction).

For a two-way interaction between features $ x_j $ and $ x_k $ for instances $ i = 1, 2, \ldots, I $, the H-statistic $ {\rm H}_{jk} $ is
\begin{equation} 
	{\rm H}_{jk}^2 ~ = ~ \dfrac{\sum_{i=1}^{I}\left[PD_{jk}\left(x_j^{\left(i\right)},x_k^{\left(i\right)}\right)-PD_j\left(x_j^{\left(i\right)}\right)-PD_k\left(x_k^{\left(i\right)}\right)\right]^2}{\sum_{i\ =1} ^{I}{PD_{jk}^2}\left(x_j^{\left(i\right)},x_k^{\left(i\right)}\right)}  
	\label{eq_hstat}
\end{equation}
where $ PD_{jk}  $ denotes the 2-way partial dependence of both $ j $ and $ k $, and $ PD_{j}  $ or $ PD_{k}  $ are 1-way partial dependence of the single features. If there is no interaction between $ x_j $ and $ x_k $, then it follows that

\begin{equation} 
	PD_{jk}\left(x_j^{\left(i\right)},x_j^{\left(k\right)}\right) ~=~ PD_j\left(x_j^{\left(i\right)}\right) + PD_k\left(x_k^{\left(i\right)}\right),
	\label{eq_pd_independent}
\end{equation}
which results in $ H_{jk} = 0$. If the two features only affect the prediction through interaction and not individually, then $ H_{jk} = 1$. 

On the other hand, the total interaction of feature $ x_j $ with all other features, $ H_j $, is
\begin{equation} 
	H_j^2 ~=~ \dfrac{\sum_{i=1}^{n}\left[\hat{f}\left(x^{(i)}\right)-PD_j\left(x_j^{\left(i\right)}\right)-PD_{-j}\left(x_{-j}^{\left(i\right)}\right)\right]^2}{\sum_{i\ =1}^{n}\widehat{f^2}\left(x^{(i)}\right)}.
	\label{eq_hstat_allinteractions}
\end{equation}
Similarly, if a feature has no interaction with any other feature, the prediction function $ \hat{f}\left(x\right)\ $ can be expressed as a sum of partial dependence functions, where the first summand depends only on $ x_j $, and the second on all other features except $ x_j $ i.e.,
\begin{equation} 
	\hat{f}\left(x\right)\ =\ PD_j\left(x_j\right)+PD_{-j}\left(x_{-j}\right),
	\label{eq_hstat_predonevsall}
\end{equation}
where $ PD_{-j}(x_{-j}) $ is the $ PD $ depending on all except $ x_j $.

\subsection{Shapley Additive Explanation}
\label{subsec_meth_shap}

SHapley Additive exPlanations (SHAP) is a local, model agnostic explainable AI (XAI) method for determining the effect of individual input variables for predicting a single product (instance) with arbitrary machine learning models, based on their complex interactions. This method uses cooperative game theory to quantify the contributions of input variables and provide a comprehensive interpretation of model predictions \cite{lundberg_unified_2017}.

The fundamental idea behind SHAP is to fairly distribute the contribution of each input variable to the model prediction. This is achieved by measuring the change in prediction when different features are included or excluded from the model. By calculating the average contribution of each feature over all possible subsets of features, SHAP values are generated that provide a fair and consistent distribution of contribution weights \cite{lundberg_unified_2017}.

The permutation-based SHAP implementation is based on the permutation version of the Shapley value equation. The machine settings are independently tuned with zero correlations among themselves, hence, the feature independence holds. 
As proposed in \cite{lundberg_unified_2017}, a sampling strategy is applied using a sample of $ P $ permutations (Equation $ \left(\ref{eq_shap1} \right)$).

Every permutation $p \in {1, 2, \dots, P}$ represents a random ordering of the input features/machine settings. 
The contribution of each input feature $ x $ is estimated for every permutation $ p $ using a weighted average as given in Equation $\left(\ref{eq_shap2}\right)$. 
A background dataset $W$ is used to represent the distribution of input features. For each instance $i$ (to be explained) and each permutation $p$, a set of ``masked" instances are created. These instances are derived from the background dataset by sequentially ``filling in" feature values from the instance $i$ according to the order specified by permutation $p$.
%
%
%
%
%
Denote 
\begin{itemize}
	\item $x_j$ as the value of $ j $-th feature in instance $ i $ that is to be explained,
	\item $O^{(p)}$ represent the ordering of features in permutation $p$, and
	\item the set of predecessors of feature $x_j$ in permutation $p$ as ${\rm Pre}^{x_j}(O^{(p)})$
\end{itemize}
For each instance $w$ in the background dataset $\mathbf{\rm W}$, create a masked instance $w'$ where
\begin{itemize}
	\item $w'_k = x_k$ if $x_k \in {\rm Pre}^{x_j}(O^{(p)}) \cup {x_j}$ (features preceding $x_j$ and $x_j$ itself take values from $ x $);
	\item $w'_k = w_k$ otherwise (other features, if any, retain their values from $w$).
\end{itemize}

The contribution of feature $x_j$ in permutation $p$ is then estimated as a weighted average of the differences in model predictions between these masked instances and instances where feature $x_j$ is also masked i.e.,
\begin{equation}
	V_{x_j, O^{(p)}}^{(i)} = \sum_{w \in \mathbf{\rm W}} \omega_p [f(w') - f(w'')],
	\label{eq_shap2}
\end{equation}
where 
\begin{itemize}
	\item $w'$ is the masked instance as defined above,
	\item $w''$ is the masked instance where $w''_k = x_k$ if $k \in {\rm Pre}^{x_j}(O^{(p)})$ and $w''_k = w_k$ otherwise (i.e., feature $x_j$ is also masked),
	\item $ V_{x_j, O^{(p)}}^{(i)} $ denotes the contribution of feature $x_j$ to the model's prediction for a specific permutation $O^{(p)}$,
	\item $ f(\cdot) $ is the model's prediction function (RF or MLP), and
	\item $\omega_j$ represents a specific weighting kernel used by Kernel SHAP \cite{lundberg_unified_2017}.
\end{itemize}

The final SHAP value for feature $x_j$, denoted as $ \phi_{x_j} $, is the average of these contributions over all $P$ permutations for instance $ i $ i.e., 
\begin{equation}
	\phi_{x_j}^{(i)} = \frac{1}{P} \sum_{p=1}^P V_{x_j, O^{(p)}}^{(i)}.
	\label{eq_shap1}
\end{equation}
%
%
Then, the sum of SHAP values for all features in instance $ i $ approximates the difference between the model's prediction of that instance and the expected model prediction over the background dataset $\mathbf{\rm W}$, which serves as the baseline. 

The SHAP and ICE explanations are called once for each production cycle of the injection moulding process. Each cycle is on the order of a few seconds (in our experiments, 10 seconds). As the explanations are called once in each cycle on the lighter models like RF and as there are 6 machine setting inputs, therefore, the SHAP's complexity, being on the order of a few milli-seconds, does not hinder the real-time deployment of explanations per each product.
%


\subsection{Cause Analysis}
\label{subsec_meth_cause_analysis}

To show that the consideration of parameter interaction alters the cause analysis in the quality assessment of injection moulded products, a machine setting's impact ranking is noted for each product sample. The rankings by ICE (the previously used interpretation method in injection moulding \cite{obregon_rule-based_2021}) are used as the benchmark for SHAP based rankings. The impact of $ j $-th feature by SHAP is the absolute of Equation $\left(\ref{eq_shap1}\right)$  i.e., $ \left|\phi_{x_j}^{(i)} \right| $, while in ICE, it is calculated by taking the standard deviation of parameters for that instance/product sample $ i $ i.e.,

\begin{equation} 
	\sigma_{x_j}^{\left(i\right)} \ = \ \sqrt{\frac{1}{C}\sum_{c=1}^{C}\left({\hat{f}}_{x_j}^{\left(i\right)}\left(c\right)-\mu\right)^2},
	\label{eq_ice_std}
\end{equation}
where $ c=1,\ldots C $ are all the possible input values of the considered feature $ x_j $ (reported in Table \ref{tab_machine_setting_values},  $ {\hat{f}}_{x_j}^{\left(i\right)}\left(c\right) $ is the ICE curve of instance $ i $ at the value $ c $, and
\begin{equation} 
	\mu=\frac{1}{C}\sum_{c=1}^{C}{{\hat{f}}_{x_j}^{\left(i\right)}\left(c\right)}.
	\label{eq_ice_mean}
\end{equation}
The higher value of $ \sigma_{x_j}^{\left(i\right)} $ means that the $ j $-th machine setting has higher impact on the product's weight in comparison to the other machine settings because the partial dependence varies more with the variation in $ x_j $. For visualization comparison of ICE and SHAP, both $ \sigma_{x_j}^{\left(i\right)} $ and $ \left|\phi^{(i)}_{x_j}\right|  $ are normalized in $ [0,\ 1] $ with 1 being the highest impact on product weight and 0 being the lowest impact. The absolute of shapely values is taken so that the feature impact is ranked irrespective of their positive or negative directions. Instead of standard deviation, the peak-to-peak difference of ICE curves may also be considered, which results in the same curves.

\section{Results and Discussion}
\label{sec_results}

All the input unique values of the dataset are reported in Table \ref{tab_machine_setting_values}. The last column represents the symbol used to denote the corresponding values in Figures \ref{fig_results_shap_1} and \ref{fig_results_shap_2}. The Pearson's correlation coefficients among the six machine settings are all zero. 

\begin{table}[h] 
	\centering
	\caption{The unique values of machine settings for the experiments with the last column denoting a short symbol for the values as used in Figures \ref{fig_results_shap_1} and \ref{fig_results_shap_2}. }
	\label{tab_machine_setting_values}
	\resizebox{\textwidth}{!}{\begin{tabular}{ccccccc}
		\hline
		\begin{tabular}[c]{@{}c@{}} Injection velocity \\ $ {\rm cm}^3/ {\rm s} $ \end{tabular} & 
		\begin{tabular}[c]{@{}c@{}} Cooling time \\ (sec) \end{tabular} &
		\begin{tabular}[c]{@{}c@{}} Packing pressure \\ (bar) \end{tabular} & 
		\begin{tabular}[c]{@{}c@{}} Packing time \\ (sec) \end{tabular} & 
		\begin{tabular}[c]{@{}c@{}} Mould temperature \\ $ \left(\ ^\circ\mathrm{C}\ \right) $ \end{tabular} & 
		\begin{tabular}[c]{@{}c@{}} Melt temperature \\ $ \left(\ ^\circ\mathrm{C}\ \right) $ \end{tabular} & Symbol \\ \hline
		23.21 & 7.79 & 550.00 & 3.03 & 60.00 & 260.00 & $ \Uparrow $ \\
		19.49 & 6.49 & 453.03 & 2.38 & 50.30 & 247.07 & $ \uparrow $ \\
		17.46 & 5.79 & 400.00 & 2.03 & 45.00 & 240.00 & $ - $ \\
		15.43 & 5.08 & 346.97 & 1.67 & 39.70 & 232.93 & $ \downarrow $ \\
		11.71 & 3.79 & 250.00 & 1.03 & 30.00 & 220.00 & $ \Downarrow $ \\ \hline
	\end{tabular}}
\end{table}

Table \ref{tab_ml_algos} compares suitable machine learning algorithms after hyperparameter optimization (HPO) through grid searching. The evaluation of seven machine learning models for predicting injection moulding parameters demonstrates that diverse algorithmic approaches can achieve high predictive accuracy. Among these, Gradient Boosting and Random Forest exhibits superior performance, achieving the lowest MAE of 0.0039 and 0.0037, and MAPE of 0.0329\% and 0.0312\% respectively. These metrics indicate errors less than 0.033\% of the actual values, which is significant in manufacturing contexts where small deviations can impact product quality.
For the explainable AI part, two high-performing yet algorithmically distinct models: Random Forest and Multilayer Perceptron (MLP) are chosen. This selection is motivated by both their predictive performance and their representativeness of different machine learning paradigms to ensure that the findings are robust to the underlying model architecture.

\begin{table}[h]
	\centering
	\caption{Error performance of suitable machine learning algorithms after HPO.}
	\label{tab_ml_algos}
	\resizebox{\textwidth}{!}{\begin{tabular}{llll}
		\toprule
		Model & MAE (HPO) & RSMAE \% (HPO) & MAPE \% (HPO) \\
		\midrule
		Gradient Boosting & 0.00386 & 0.7877 & 0.0329 \\
		XGBoost & 0.009073 & 1.8517 & 0.0773 \\
		Support Vector Regressor & 0.018308 & 3.7363 & 0.1559 \\
		K-Nearest Neighbors & 0.007922 & 1.6167 & 0.0675 \\
		Ridge Regression & 0.005593 & 1.1414 & 0.0476 \\
		MLP & 0.0051 & 1.03 & 0.0431 \\
		Random forest & 0.0037 & 0.79 & 0.0327 \\
		\bottomrule
	\end{tabular}}
\end{table}

Table \ref{tab_mae} reports the error performance of both RF and MLP on the unseen test set. The models' performances are deemed acceptable to proceed with the model's interpretations because the error with respect to the scale of the data (highlighted by MAPE) and with respect to the dynamic range of the data (indicated by RSMAE) are acceptable.

The MLP model's training stability is rigorously evaluated through systematic analysis of 10 independent and random runs with varied initializations, revealing expected sensitivity to initialization conditions, a well-documented challenge in neural network optimization \cite{bengio_learning_1994}. While 80\% of runs achieve stable convergence (MAE: 0.0051–0.0078), 20\% exhibited premature stagnation (MAE $ \approx $ 0.086). The MLP model exhibited initialization-sensitive convergence behaviour consistent with neural network optimization challenges. 
This approach achieves robust test performance (MAE = 0.0051) while maintaining computational efficiency, with architectural constraints (8,4 network) mirroring successful implementations that prioritizes stability over complexity escalation.

\begin{table}[h]
	\centering
	\caption{Error performance (95\% confidence interval (CI)) of RF and MLP as measured by MAE, MAPE, and RSMAE}
	\label{tab_mae}
	\resizebox{\textwidth}{!}{\begin{tabular}{llll}
		\hline
		Model & MAE (95\% CI) & MAPE (95\% CI) & RSMAE (95\% CI) \\ \hline
		RF & 0.0037 (0.0036, 0.0037) & 0.0312\% (0.0308, 0.0315)\% & 0.79\% (0.7792, 0.8021)\% \\
		MLP & 0.0051 (0.0048, 0.0052) & 0.0431\% (0.0425, 0.0434)\% & 1.03\% (0.99, 1.06)\% \\ \hline
	\end{tabular}}%
\end{table}

\subsection{Feature Interactions}
\label{subsec_results_feature_interact}

\begin{figure}[h]
	\centering
	\includegraphics[scale=0.3]{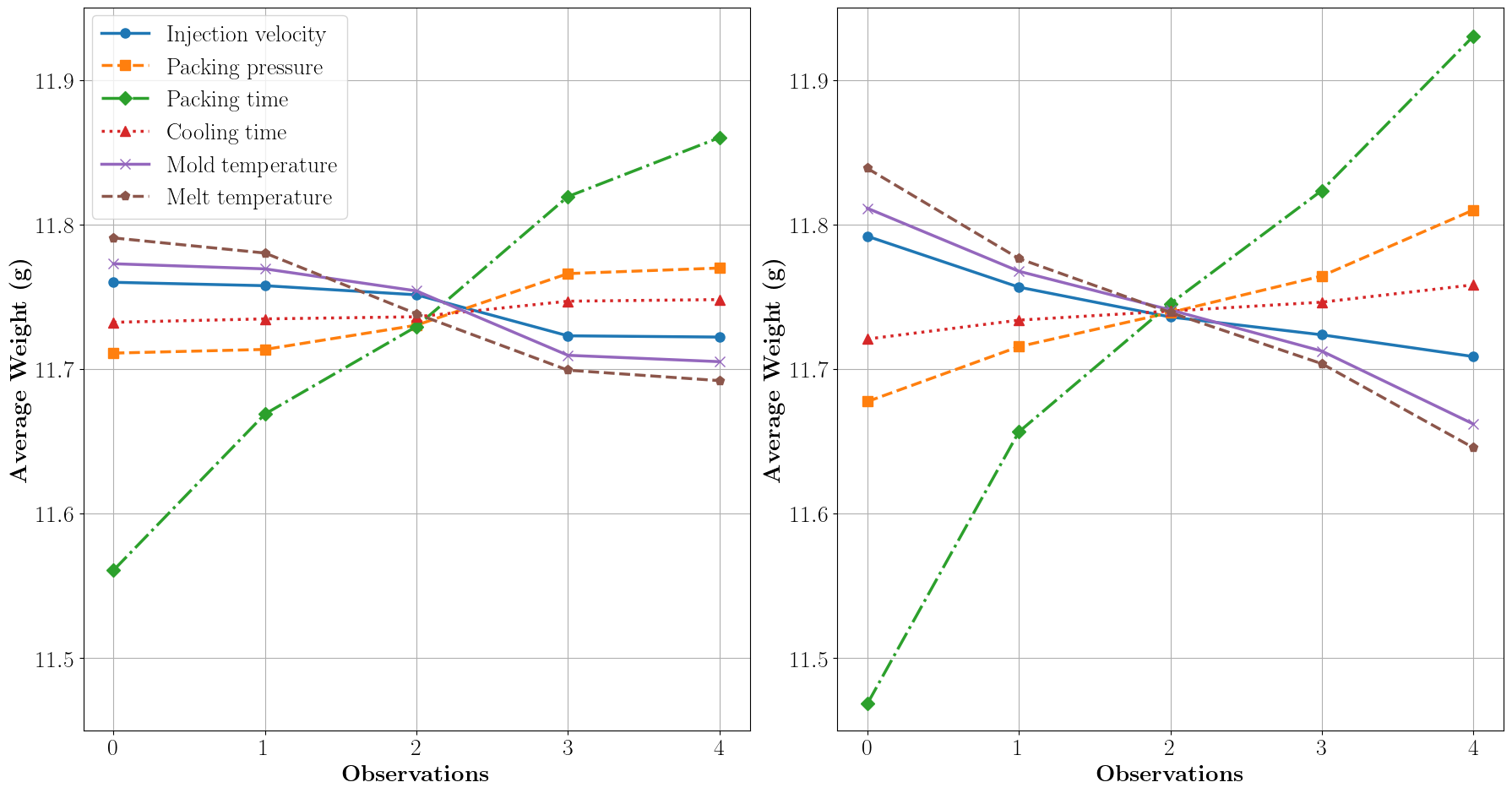}
	\caption{PDP plots of machine settings for (a) RF, and (b) MLP.} 
	\label{fig_res_pdp}
\end{figure}

The trained RF and MLP models are fed to the PDP explainer to analyse the feature impacts on the product weight and are shown in Figure \ref{fig_res_pdp} for all unique values of the machine settings. The units and scale of these machine settings are different, and they are plotted together merely to show their partial dependence. The results are consistent across the two models, with both ranking packing time as the most important feature, followed by the melt temperature. None of them shows flat line, which indicates that there may exist interactions among the machine settings in impacting the product weight. 

\begin{figure}[h] 
	\centering
	\includegraphics[scale=0.45]{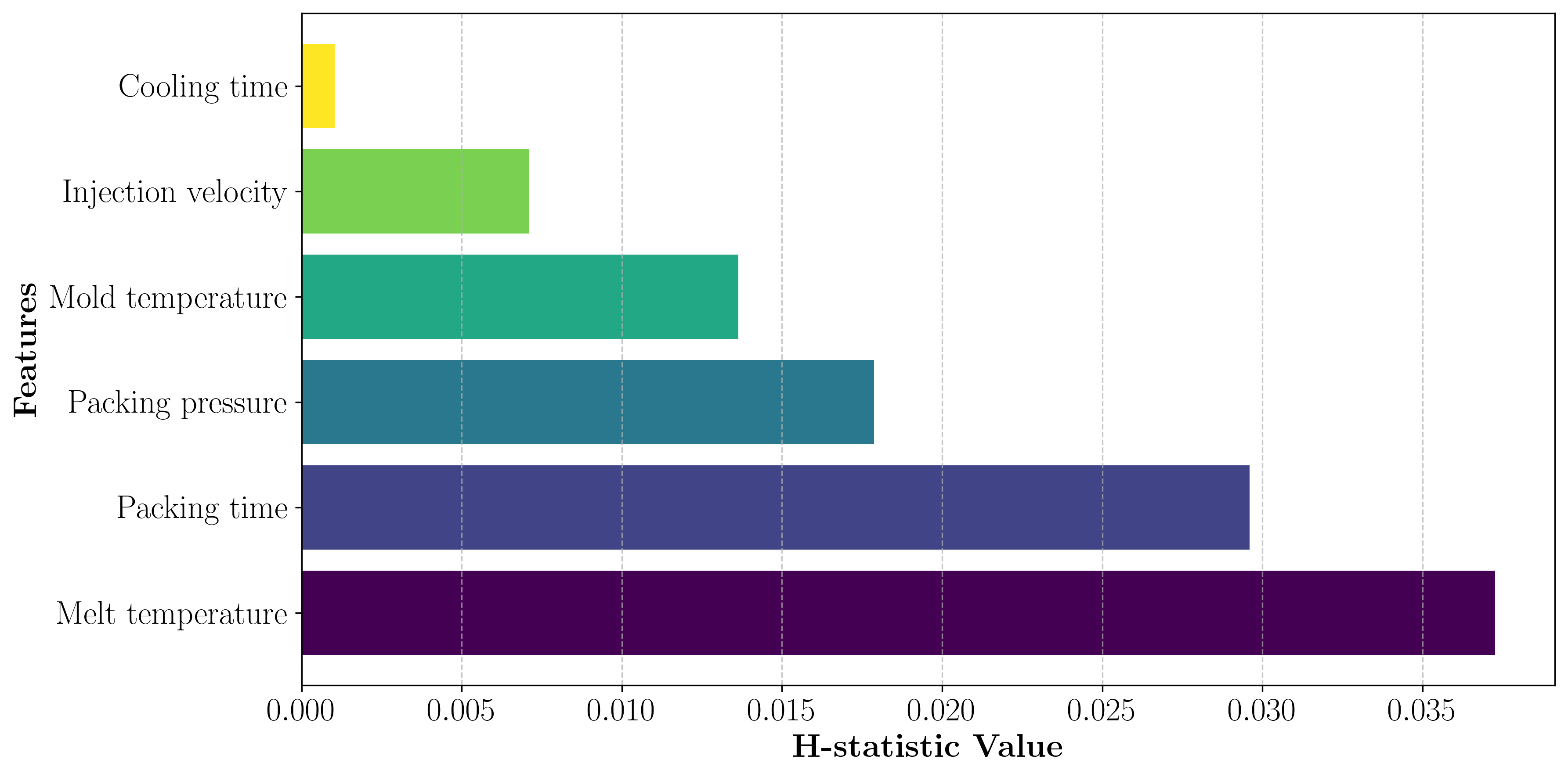}
	\caption{Bar plot of one-versus-all interactions of machine settings measured by H-statistic based on RF.}
	\label{fig_res_hstat}
\end{figure}

To show the feature interactions, 100 random instances from the test set are used to calculate the H-statistic (Figure \ref{fig_res_hstat}). 
For brevity, only the RF model is used.
There are interactions among the features, with melt temperature and packing time accounting for maximum of them. These interactions may or may not be significant, which can be checked by varying a feature while keeping another constant and plotting them against the product weight. This is described next.

\begin{figure}[h] 
	\centering
	\includegraphics[scale=0.26]{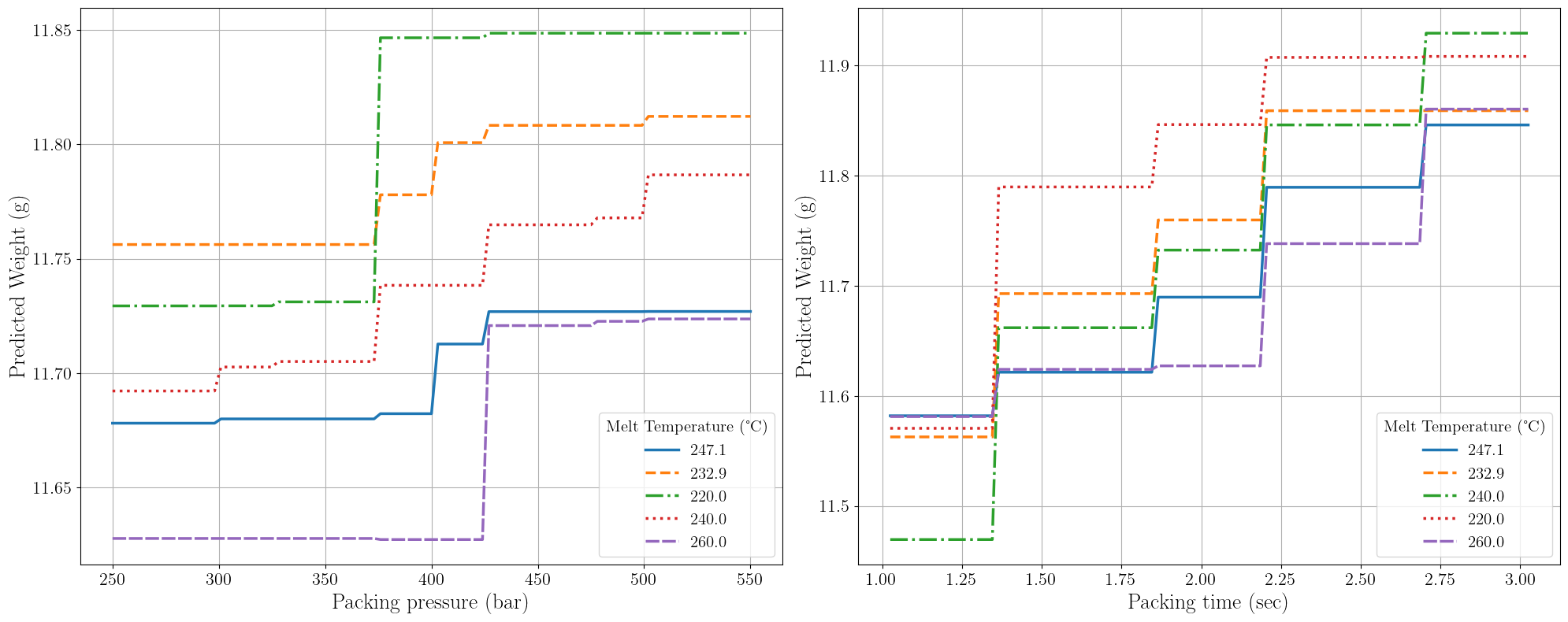}
	\caption{Step plots of packing time (left) and packing pressure (right) vs RF-based weight predictions for different values of the melt temperature.}
	\label{fig_res_step_plots}
\end{figure}

Let's look at the plots of a feature vs predicted product weight with respect to another feature for which the H-statistic shows more interactions i.e., melt temperature, packing time, and packing pressure (Figure \ref{fig_res_step_plots}) \cite{molnar_interpretable_2022}.
The model predicts on 100 uniformly-sampled values in the dynamic range of the ordinates (packing time and packing pressure). 
The remaining machine settings are held constant at their averages. At different values of melt temperature, the packing pressure and packing time curves behave differently indicating interactions. For instance, 
when switching the packing pressure from 346.97 bar to 453.03 bar, the weight change is 0.12 g when the melt temperature is set at $ {220}^\circ\mathrm{C} $ and the weight change is 0.05 g when the melt temperature is set at $ {247.1}^\circ\mathrm{C} $, which is half as much change as the former setting.

Similarly, in Figure \ref{fig_res_step_plots} (right), the peak-to-peak difference of weight prediction is maximum for the melt temperature curve of $ {240}^\circ\mathrm{C} $ indicating the maximum impact by packing time i.e., when switching packing time from 1.03 sec to 3.03 sec: for the melt temperature being $ {240}^\circ\mathrm{C} $ vs it being $ {247}^\circ\mathrm{C} $, the weight change is 0.49 g vs 0.25 g, which is again a factor of 2 difference. These observations indicate that the interactions of melt temperature with packing pressure and packing time are significant in impacting the product weight.

\begin{figure}[h!] 
	\centering
	\includegraphics[width=\textwidth]{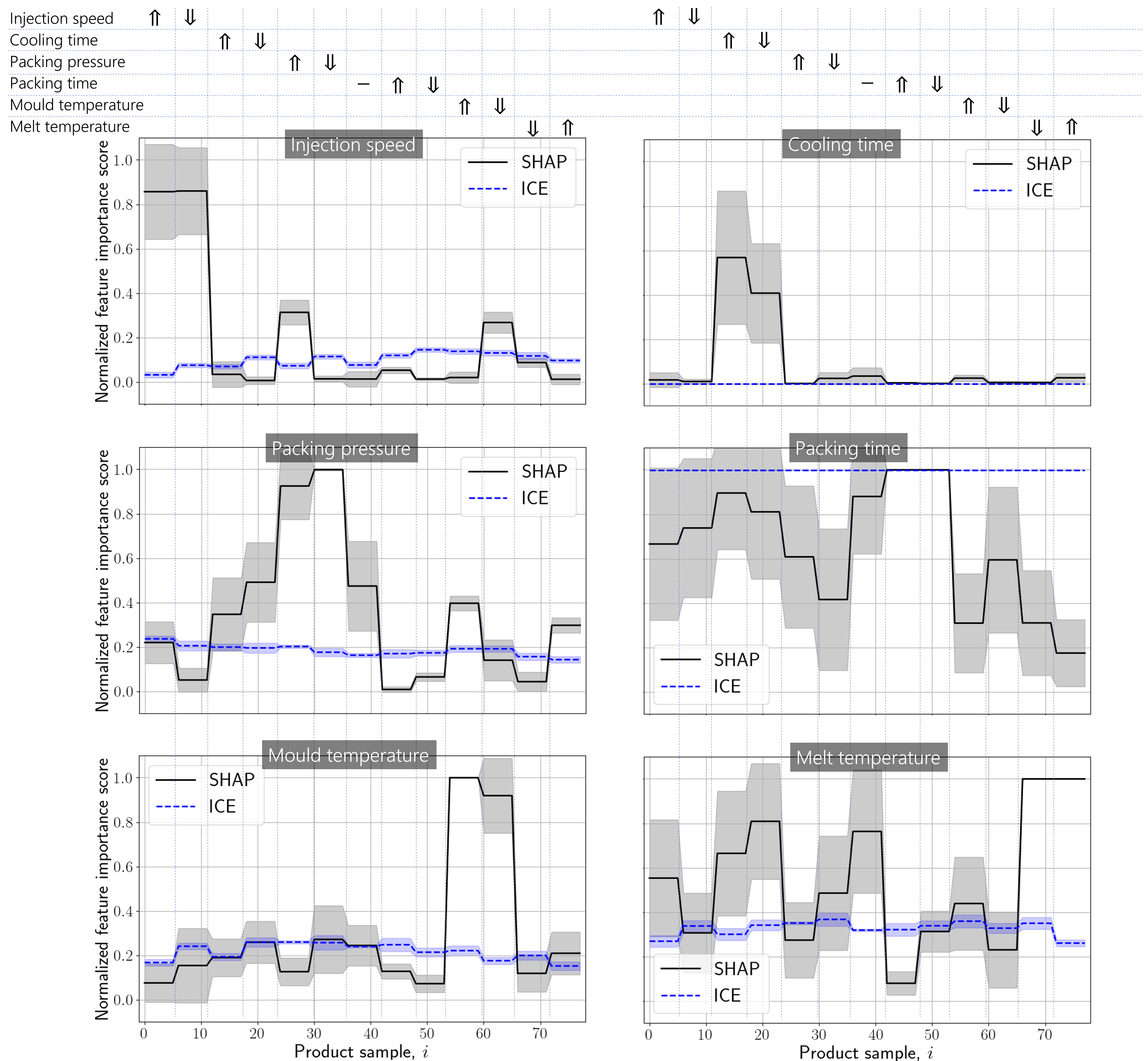}
	\caption{Comparison of the normalized feature impact values of SHAP and ICE applied on RF for all machine settings. Here, $ \Uparrow $ indicates the highest experimented value of the machine setting (Table \ref{tab_machine_setting_values}), $ \Downarrow $ denotes the lowest experimented value, and $ – $ or symbol's absence represents the mid value. The curve shades denote the standard deviation from the random trials.}
	\label{fig_results_shap_1}
\end{figure}

\begin{figure}[h!] 
	\centering
	\includegraphics[width=\textwidth]{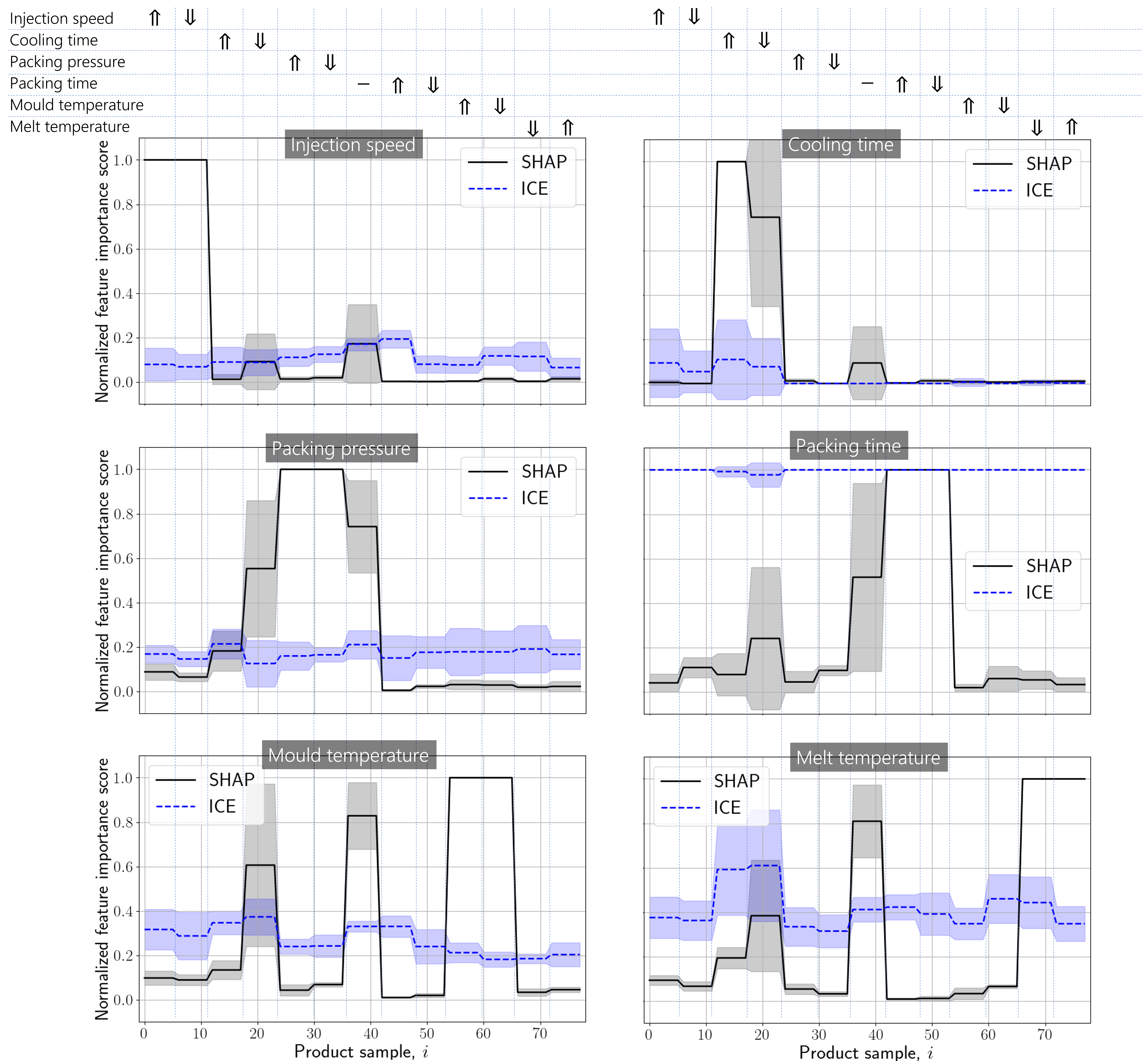}
	\caption{Comparison of the normalized feature impact values of SHAP and ICE applied on MLP for all machine settings. Here, $ \Uparrow $ indicates the highest experimented value of the machine setting (Table \ref{tab_machine_setting_values}), $ \Downarrow $ denotes the lowest experimented value, and $ – $  or symbol's absence represents the mid value. The curve shades denote the standard deviation from the random trials.}
	\label{fig_results_shap_2}
\end{figure}

Plots of the impact rankings of individual machine settings (described in Section \ref{subsec_meth_cause_analysis}) calculated by both SHAP and ICE are shown for the test set in Figure \ref{fig_results_shap_1} (applied on RF) and Figure \ref{fig_results_shap_2} (applied on MLP). The product weight is ideal when all the machine settings are at their mid values (see Table \ref{tab_machine_setting_values}). For each machine setting combination, 6 product samples (cycles) exist in the test set and plotted adding to the statistical rigor of the findings. The ordinates of the two curves are the normalized $ \sigma_i^{\left(x\right)} $ (for ICE) and $ \left|\phi^{(i)}_{x_j}\right|  $ (for SHAP), while the abscissa constitutes the product sample indices $ i $. For further statistical rigor, 10 repeated trials are conducted with random data splits. The mean and standard deviation of the 10 trials are calculated and visualized by the plots. On top of each figure, the corresponding 6 machine setting values of product sample $ i $ are indicated for ease of visualization by their symbols (as mapped in Table \ref{tab_machine_setting_values}).

It is observed that due to the different impact rankings, the interpretations of ICE and SHAP are quite different over the common model (may it be RF or MLP). For instance,

\begin{enumerate}
	\item[(i)]Packing time is the most affecting machine setting feature as per ICE for all product samples suggesting that it always has the most influence on the value of the product weight and that packing time is to be tuned for changing the weight no matter which other machine setting is altered. On the other hand, SHAP suggests that only when the packing time is increased/decreased, then it is the parameter that should be tuned first; otherwise, the other machine settings are suggested first to be tuned in modifying the product weight to the desired (expected) value. 
	\item[(ii)] For all machine settings (apart from the cooling time in RF), SHAP suggests tuning that machine setting first which is modified. This shows that SHAP has followed the correct input/output relationship and implies that SHAP will suggest the correct parameter tuning to achieve the expected outcome of the product weight. On the other hand, ICE will always suggest tuning the packing time, no matter which of the six machine settings is altered. 
\end{enumerate}

There are slight differences in the interpretations when SHAP or ICE is implemented on RF and MLP, but they are trivial. Such differences may attribute to the different relationships the models detect between the inputs and the output, the standardization of input data only in the case of MLP (but not RF), and the non-zero mean absolute errors.

\subsection{Feature Attribution for Cause Analysis}
\label{subsec_results_attribution}

To judge the correct feature (root cause) of the deterioration of a product's quality, \textit{controlled experiments} are performed in which a single feature is changed from its mid value and explanations are checked for their correctness. The mid values of the machine settings (Table \ref{tab_machine_setting_values}) provide the desired/expected product weight assumed in the experiments. When a single machine setting changes from its mid value causing a deviation of the product weight from its expected/desired value, it is obvious that that particular machine setting should be the first major cause of the deviation and should be attributed for the aberration. The data reflects this behaviour and a complex enough machine learning model, which is fitting that data, will also reflect this. As such an accurate model is often a black box, therefore, the interpretation method enlightening this black box model for us should also reflect this behaviour as is the case in SHAP in Figures \ref{fig_results_shap_1} and \ref{fig_results_shap_2}. However, in the case of the machine setting rankings by ICE, packing time is always suggested to be the first major cause, no matter if we have changed any of the six machine settings. Suggesting to tune the packing time always is not the wrong move if the product weight needs to be changed but packing time may not be the main cause of the weight deviation and if deployed in a real setup, an operator (human or machine) might be directed to cater for the wrong major cause of the weight change. This wrong cause analysis by ICE may aggravate the situation especially when multiple output quality characteristics (in addition to the product weight) are modelled simultaneously because even if ICE is correct in getting back the desired product weight, due to the wrong major cause identification, it is probable that other quality characteristics are adversely affected. This suggests that in the quality prediction of the designed experiment, SHAP is the viable option to identify the correct cause of deviation of the product's weight in comparison to ICE. 

\begin{figure}[h!] 
	\centering
	\includegraphics[scale=0.2]{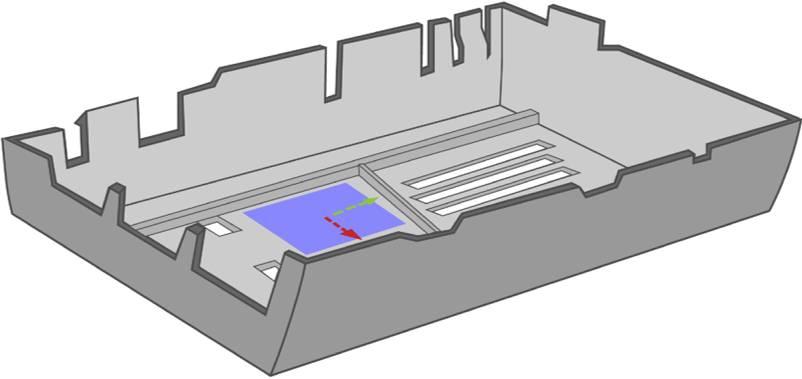}
	\caption{Product sample showing the shaded region of measuring the planarity characteristic.}
	\label{fig_planarity}
\end{figure}

\begin{figure}[h!] 
	\centering
	\includegraphics[width=\textwidth]{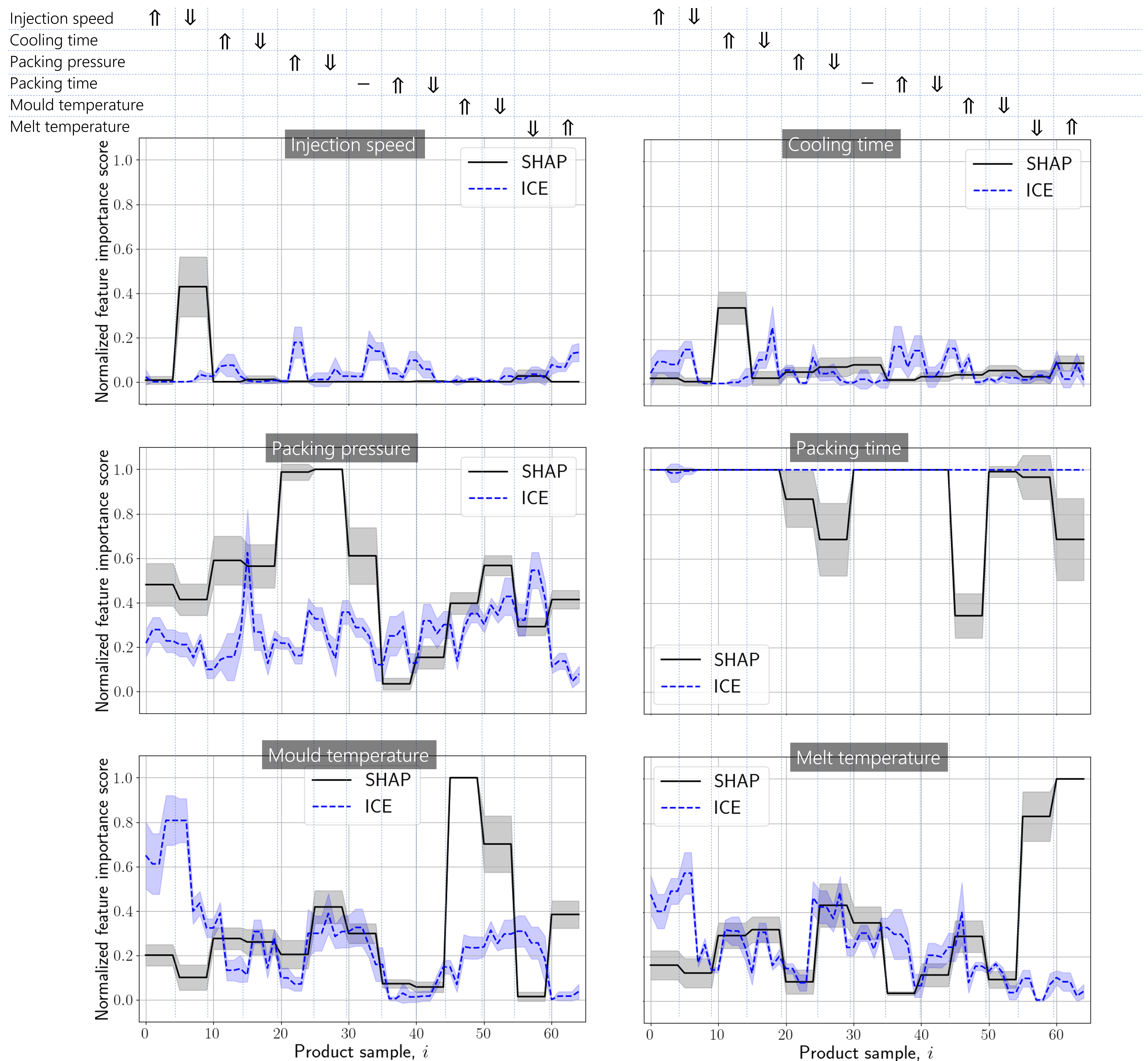}
	\caption{Comparison of the normalized feature impact values of SHAP and ICE applied on RF for all machine settings for product's planarity of the shaded area. Here, $ \Uparrow $ indicates the highest experimented value of the machine setting (Table \ref{tab_machine_setting_values}), $ \Downarrow $ denotes the lowest experimented value, and $ – $  or symbol's absence represents the mid value. The curve shades denote the standard deviation from the random trials.}
	\label{fig_results_shap_3}
\end{figure}

This should hold true for other production setups in injection moulding too because unlike ICE, SHAP's impact assessment of a machine setting/feature involves the contribution of other machine settings/features, and in injection moulding, features do interact in impacting a product's characteristic as shown in this study by the H-statistic (Figure \ref{fig_res_hstat}) as well as indicated in some previous studies \cite{obregon_rule-based_2021, knollAnalysisMachineSpecificBehavior2023}. 
Nonetheless, in order to validate the findings in another setup, the same RF modelling pipeline and explainability methods are implemented in a different scenario: The machine settings are now attributed to a product quality characteristic measuring planarity/flatness/levelness along the shaded region of the product as shown in Figure \ref{fig_planarity}. The normalized $ \sigma_i^{\left(x\right)} $ (for ICE) and $ \left|\phi^{(i)}_{x_j}\right| $ (for SHAP) are plotted against the product samples, indexed as $ i $, and shown in Figure \ref{fig_results_shap_3}.
Again, the product planarity is ideal when all the machine settings are at their mid values (see Table \ref{tab_machine_setting_values}). On top of each figure, the corresponding 6 machine setting values of product sample $ i $ are indicated for ease of visualization by their symbols (as mapped in Table \ref{tab_machine_setting_values}).
When a machine setting is changed from its ideal (mid) value, $ \sigma_i^{\left(x\right)} $ keeps the packing time to be the major cause, while $ \left|\phi^{(i)}_{x_j}\right| $ increases in favour of the machine setting that is varied. Therefore, for ICE, packing time still remains the main cause for any experimental cycle, while for SHAP, it varies in favour of the actual cause of deviation from the ideal settings.

\section{Conclusion}
\label{sec_concl}

It is observed that the important uncorrelated machine settings in injection moulding (e.g., melt temperature and packing time for product's weight in the experiment of this study) do interact to impact the weight of the output product as shown by the H-statistic. These interactions are captured by both the permutation-based SHAP and the previously employed ICE methods. For a new product cycle, identifying the cause of an undesired product quality is achieved by ranking the impact of machine settings using these local model interpretations. However, the proposed permutation-based SHAP performs accurate cause identification when a product deviates from its expected weight (as shown in Figures \ref{fig_results_shap_1} (for random forest) and \ref{fig_results_shap_2} (for multilayer perceptron)). The correct cause identification leads to a more reliable and quality-consistent production in an automated mass production setup.
Furthermore, it will be interesting to explore how the cause analysis/feature attribution extends to the case of multiple product quality characteristics (particularly the defects) because in such cases, if the correct cause is not identified, the other quality characteristics can be deteriorated that may impose further challenges in achieving the desired machine settings.

\section{Acknowledgment}
\label{acknowl}

This project is funded by AIR@InnoHK research cluster of the Innovation and Technology Commission (ITC), Hong Kong. 
We like to acknowledge Smart MM Engineering Systems Limited, Dongguan, China, particularly Professor Fred Law and Dr. Wing Hong Szeto, for sharing their valuable insights from practical industrial experiences.  
We would also like to thank the Institute for Plastics Processing (IKV) in Industry and Craft at RWTH Aachen University, Germany for their collaboration on machine setup and production data collection.



\bibliographystyle{elsarticle-num} 
\bibliography{AIQP2_refs}


\end{document}